\documentclass[sigconf,]{acmart}
\setcopyright{none}
\usepackage{multirow}
\usepackage{amsthm}
\newtheorem{theorem}{Theorem}
\newtheorem{definition}{Definition}
\usepackage[normalem]{ulem}
\useunder{\uline}{\ul}{}
\AtBeginDocument{%
  \providecommand\BibTeX{{%
    \normalfont B\kern-0.5em{\scshape i\kern-0.25em b}\kern-0.8em\TeX}}}

\acmConference[Conference acronym 'XX]{Make sure to enter the correct
  conference title from your rights confirmation emai}{June 03--05,
  2018}{Woodstock, NY}
\begin{document}

\title{GFairHint: Improving Individual Fairness for Graph Neural Networks via Fairness Hint}






\author{Paiheng Xu $^{* \ 1}$, Yuhang Zhou $^{* \ 2}$, Bang An $^{1}$, Wei Ai $^{2}$, Furong Huang $^{1}$}
\affiliation{\institution{$^{1}$ Department of Computer Science, University of Maryland, College Park}}
\affiliation{\institution{$^{2}$ College of Information Studies, University of Maryland, College Park}}
\email{{paiheng, tonyzhou, bangan, aiwei, furongh}@umd.edu}

\renewcommand{\shortauthors}{Paiheng Xu et al.}

\begin{abstract}
Given the growing concerns about fairness in machine learning and the impressive performance of Graph Neural Networks (GNNs) on graph data learning,  algorithmic fairness in GNNs has attracted significant attention. 
While many existing studies improve fairness at the group level, only a few works promote individual fairness, which renders similar outcomes for similar individuals. 
A desirable framework that promotes individual fairness should (1) balance between fairness and performance, (2) accommodate two commonly-used individual similarity measures (externally annotated and computed from input features), (3) generalize across various GNN models, and (4) be computationally efficient.
Unfortunately, none of the prior work achieves all the desirables.
In this work, we propose a novel method, \textit{GFairHint}, which promotes individual fairness in GNNs and achieves all aforementioned desirables.
GFairHint learns fairness representations through an auxiliary link prediction task, and then concatenates the representations with the learned node embeddings in original GNNs as a \emph{``fairness hint''}.
Through extensive experimental investigations on five real-world graph datasets under three prevalent GNN models covering both individual similarity measures above, GFairHint achieves the best fairness results in almost all combinations of datasets with various backbone models, while generating comparable utility results, with much less computational cost compared to the previous state-of-the-art (SoTA) method. 

\end{abstract}

\maketitle

{\let\thefootnote\relax\footnotetext{{$^*$ equal contribution}}}
\section{Introduction}
\label{sec:intro}
Graph Neural Networks (GNNs) have shown great potential in modeling graph-structured data for various tasks such as node classification, graph classification, and link prediction \cite{wu2021comprehensive}. 
Node classification, particularly, has been applied in many real-world scenarios, such as recruitment \cite{lambrecht2019algorithmic, zhang2021attentive}, recommendation system \cite{li2020hierarchical, 10.1145/3535101, 10.1145/3308558.3313442,wu2022graph}, and loan default prediction \cite{hu2020loan, wang2022review}. 
Although, GNNs play important roles in these decision-making processes, increasing attention is being paid to fairness in graph-structured data and GNNs \cite{10.1145/3457607, dong2022fairness, lambrecht2019algorithmic}.
Since the node representation is learned by aggregating information from its neighbours, the unfairness may be amplified due to such message-passing mechanism in GNNs and result in unexpected discrimination \cite{dong2022structural, wang2022fairview, kose2022fair}. 
Taking social networks as an example, users tend to connect with others in the same demographic group. 
The message-passing mechanism may cause GNNs to perform differently for different demographic groups and have different predictions for similar individuals if they belong to different demographic groups. 
Such unfairness prevents the adoption of the GNNs in many high-stake applications.

Algorithmic fairness can be defined at both group and individual level~\cite{mehrabi2021survey}. 
While group fairness attempts to treat different groups equally, \emph{individual fairness}, which is the focus of our work, intends to \textit{give similar predictions to similar individuals} for a specific task. Compared to group fairness, individual fairness reduces bias directly on each individual. 
In particular, when information on sensitive attributes is not available in real-world applications, we can only choose individual fairness to scrutinize the discrimination.


A core question for individual fairness is how to define the task-specific similarity metric that measures how similar two individuals are. 
\citet{dwork2012fairness} originally envisioned that the metric would be provided by human experts ``as a (near ground-truth) approximation agreed upon by the society''. 
\citet{lahoti2019operationalizing} argues that it is very difficult for experts to measure individuals based on a quantitative similarity metric when in operationalization.
They further suggest that it is much easier to make pairwise judgments which results in a binary similarity measure between two individuals.
Class-specific similarity where individuals in the same class are considered similar can also be transformed in the binary format.
For cases where there is no task-specific similarity metric at hand, other works \cite{10.5555/3524938.3525596, dong2021individual, kang2020inform} use simplified notions by computing continuous similarity metrics (e.g., a weighted Euclidean distance) over a feature space of data attributes. Thus, we argue that an ideal framework to promote individual fairness in models should be compatible with multiple task-specific similarity metrics.


\begin{figure*}[!ht]
    \centering
    \includegraphics[width=0.9\linewidth]{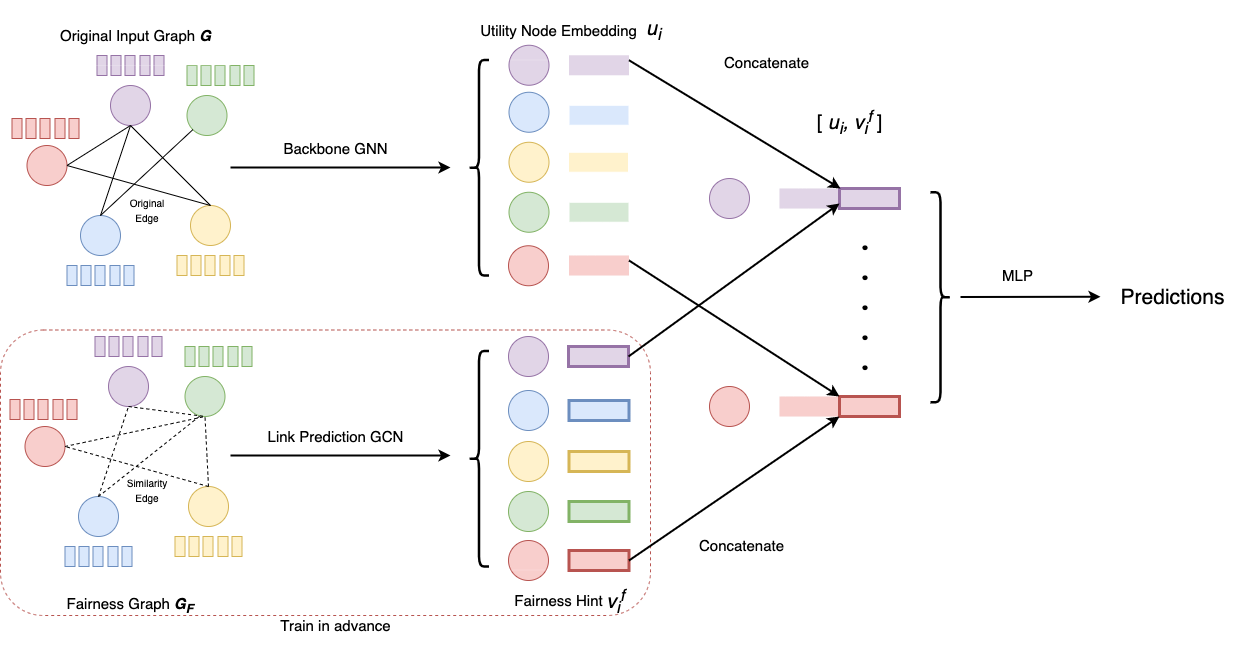}
    \vspace{-1em}
    \caption{The proposed individual fairness promotion framework, \textbf{GFairHint}. Colored rectangles denote the representations of corresponding nodes (i.e., individuals). GFairHint learns the fairness hint from the fairness graph and concatenates it with the utility node embedding from the original backbone GNN model. Finally, it feeds the concatenated node embedding into an MLP to make fair predictions. The loss function for GFairHint can be a single utility loss (cross-entropy loss) or a combination of utility loss and fairness loss (e.g., ranking-based loss).}
    \label{fig:model_diagram}
\end{figure*}

We highlight four desiderata for prompting individual fairness in GNNs based on the discussion above and practical experience when adding a fairness promotion module to the models. 
(1) The model should achieve a good balance between utility (e.g., classification accuracy) and fairness when making predictions. 
(2) The proposed method for individual fairness should be compatible with both binary and continuous similarity measures.
(3) The fairness promotion method should be compatible with different GNN model architectures and various designs for specific tasks.
(4) The additional computational cost introduced to promote fairness should be reasonably small. 


In this study, we propose an individual fairness representation learning framework to improve individual \underline{FAIR}ness for \underline{G}NNs via fairness \underline{HINT} (\textbf{GFairHint}), with the above desired properties. 
We consider the setting in which similarity measures are available for each pair of individuals, either binary or continuous.
As shown in Figure \ref{fig:model_diagram}, in addition to the original input graph, we create a \textit{fairness graph} where the edge between two nodes is weighted by the given similarity measure and does not exist when the similarity value is $0$ or below a certain threshold.
We then learn a fairness representation for each node from the constructed fairness graph via link prediction, where we encourage the model to recover randomly masked edges.
The learned fairness representation is then used as \textit{fairness hint} to be concatenated with the node embeddings from the original graph, which is trained in parallel to maximize utility with the main GNN model.
We feed the concatenated representations to multilayer perceptrons (MLPs) to make fair and accurate predictions.


GFairHint focuses on learning an additional fair representation and is compatible with various GNN model designs. 
It is orthogonal and complementary to a strategy adopted by most previous work \cite{lahoti2019operationalizing,dong2021individual,kang2020inform}, i.e., adding a fairness regularization term to the training objective. 
Meanwhile, GFairHint is a light-weighted framework. 
It is more computationally efficient and especially works better with large network datasets in terms of computation cost and both utility and fairness performance, while previous work either takes much longer training time \cite{dong2021individual} or requires large memory allocation, and therefore may fail on some large datasets \cite{lahoti2019operationalizing}.
Furthermore, GFairHint can benefit from these existing methods with fairness regularized objective functions when integrated together to further improve the performance.

To demonstrate the effectiveness of our proposed method, we conduct extensive empirical evaluations on five datasets for node classification. These datasets apply either continuous similarity measures derived from the input space or binary measures provided by external annotators to quantify the similarity between individuals. Additionally, we experiment with three popular GNN backbone models, resulting in a total of 15 models $\times$ dataset comparisons.
Our GFairHint framework consistently outperforms other methods in terms of fairness, achieving the best results in 12 out of the 15 comparisons. Furthermore, in 9 out of the 15 comparisons, GFairHint achieves the best utility performance, while in the remaining comparisons, it achieves comparable utility results.
We summarize our main contributions as follows:
\begin{itemize}
    \item \textbf{Plug-and-play Framework}: We present GFairHint, a plug-and-play framework for enhancing individual fairness in GNNs. This framework learns a fairness hint through an auxiliary link prediction task. We provide a theoretical analysis on the fairness hint and prove that for any two nodes, the learned hints are individually fair.
    \item \textbf{Satisfied Desiderata}: GFairHint is compatible with two distinct settings for similarity measures among individuals and can achieve comparable accuracy while generating more individually fair predictions. Additionally, the method is computationally efficient and seamlessly integrates with various model designs.
    \item \textbf{Rigorous Experiments}: We empirically show that the proposed method achieves the best fairness results in most comparisons (12/15), with the best utility results in the 9/15 comparisons, and comparable utility performance in the other comparisons.
\end{itemize}

\section{Related Work}
\paragraph{Fairness for Graph-structured Data}
Most previous efforts focus on promoting group fairness in graphs \cite{ijcai2019p456, li2021dyadic, wang2022fairview, bose2019compositional, dai2021say, pmlr-v119-buyl20a}, which encourages the same results across different sensitive groups (e.g., demographics). 
Another line of research work is on counterfactual fairness \cite{kusner2017counterfactual, ma2022learning}, which aims to generate the same prediction results for each individual and its counterfactuals. 

Few research studies individual fairness in graphs \cite{dwork2012fairness, mehrabi2021survey}. 
Individual fairness intends to render similar predictions to similar individuals for a specific task.  
\citet{kang2020inform} propose a framework called InFoRM (Individual Fairness on Graph Mining) to debias a graph mining pipeline from the input graph (preprocessing), the mining model (processing), and the mining results (postprocessing), but not specifically for GNN models.
\citet{song2022guide} identify a new challenge to enforce individual fairness informed by group equality. Promoting group and individual fairness at the same time requires group information, which is not available in some real-life scenarios, such as the academic networks studied in this paper.
The work closest to ours is REDRESS \cite{dong2021individual}. They propose to model individual fairness from a ranking-based perspective and design a ranking-based loss accordingly. 
However, their method does not generalize well when the similarity measure is binary, especially binary, because calculating the ranking-based loss requires to rank the individuals based on the similarity values but with binary similarity measures (i.e., many individuals on the same similarity level) the rankings are not as informative as in the cases with continuous similarity measures. 
Moreover, despite their effort on reducing the computation cost and the effectiveness of the ranking-based loss, high computational cost is unavoidable when computing the rank.

\paragraph{Individual Fairness}
There are other works focusing on individual fairness, but not specifically for graph-structured data. The definition of individual fairness, similar predictions for similar individuals, can be formulated by the Lipschitz constraint, which inspires works such as Pairwise Fair Representation (PFR) \cite{lahoti2019operationalizing} to learn fair representation as input. This can be considered as a preprocssing method for GNN models. However, the transformation from the original input feature to fair representation may distort the original information in the input features and cause a detriment in the utility performance of the model \cite{dong2021individual}.
Because it is computationally difficult to enforce Lipschitz constraint, \citet{yurochkin2021sensei} propose an in-process method with a lifted constraint and \citet{NEURIPS2021_d9fea4ca} propose a post-processing method with Laplacian smoothing. 
In this work, we compare with REDRESS \cite{dong2021individual} and adapt PFR \cite{lahoti2019operationalizing} and InFoRM \cite{kang2020inform} to work with GNN models. We do not compare with \citet{dwork2012fairness} because their contribution is mainly conceptual and it was not included as baseline in previous works \cite{lahoti2019operationalizing, kang2020inform, dong2021individual}.

\section{Proposed Method - GFairHint}

\subsection{Problem Formulation}
The generic definition of individual fairness is \textit{individuals who are similar should have similar outcomes} \cite{dwork2012fairness}. 
We can formulate the similarity between individuals with an \textbf{oracle similarity matrix} $\mathcal{S}_F$,
where the value of $(i, j)$-th entry is the similarity between the node $i$ and $j$.
We follow the same setting from previous works \cite{dong2021individual, kang2020inform, lahoti2019operationalizing}, 
where the oracle similarity matrix $\mathcal{S}_F$ is given apriori (annotated by external experts or calculated by input features). Depending on the definition of similarity measure, the entry of $\mathcal{S}_F$ could be continuous or binary. 
For graph data and GNN models,
we denote the \textbf{outcome similarity matrix} as $\mathcal{S}_{\hat{Y}}$ with the predicted outcome $\hat{Y}$ from GNN model, where the $(i, j)$-th entry is the similarity between the embeddings $z_i$ and $z_j$ of the nodes $i$ and $j$ from the last model layer.

Inspired by the individual fairness definition from the previous work \cite{dwork2012fairness,kang2020inform}, we define the fair condition of GNN model as follows:
\begin{definition}
\label{def:fair}
Let $x_i$ and $x_j$ be two nodes in a graph. The output of the model $f(x_i)$ and $f(x_j)$ are \textbf{individually fair} w.r.t. to the node similarity measure $S$ and output distance measure $D$ if the following condition holds.
\begin{equation}
    D(f(x_i), f(x_j))  \leq \frac{\epsilon}{S(x_i, x_j)} \forall x, y = 1,..., n
\end{equation}
where $\epsilon$ > 0 is a constant for fairness tolerance and $0 < S(x_i, x_j) < 1$ if $i \neq j$. $n$ is the number of nodes in a graph. We define $\frac{\epsilon}{S(x_i, x_j)}$ as the \textbf{fairness bound}.
\label{definition}
\end{definition}
With this definition, our goal to promote the individual fairness can be achieved by narrowing the difference between $\mathcal{S}_F$ and $\mathcal{S}_{\hat{Y}}$.

\subsection{Overall Structure}

To achieve a good balance between utility and fairness,
GNN models need to utilize input information from both sides.
We first apply a representation learning method to extract the fairness information from the input and add the learned fairness representation to the original GNN model to promote individual fairness.
Our proposed \textbf{GFairHint} framework, as shown in Figure \ref{fig:model_diagram}, consists of three steps. First (Section~\ref{sec:fair-graph}), we construct an unweighted fairness graph, $\mathcal{G}_F$, with the same set of nodes in the original input graph. The undirected edges of $\mathcal{G}_F$ represent that two nodes have a high similarity value in $\mathcal{S}_F$. 
Next (Section~\ref{sec:fair-rep}), we obtain the individual fairness hint through a representation learning method that learns fairness representations for the nodes in $\mathcal{G}_F$. Specifically, the representation learning model predicts whether two nodes in $\mathcal{G}_F$ have an edge through a GNN link prediction model whose final hidden layer output is used as the fairness hint. Finally (Section~\ref{sec:fairness_promotion}), we concatenate the node fairness hint with the learned node embedding of the GNN for original tasks and use the joint embedding for further training.

\subsection{Construction of Fairness Graph}
\label{sec:fair-graph}


To extract fairness information for each individual via a link-prediction-based representation learning method, we first construct a fairness graph $\mathcal{G}_F$ based on the apriori oracle similarity matrix $\mathcal{S}_F$.
Note that $\mathcal{S}_F$  can be given by incorporating various types of data sources and our fairness graph construction complies with the data sources of $\mathcal{S}_F$. Here, we show two commonly utilized data sources, i.e., external annotation and input feature.

\paragraph{\textbf{Oracle Similarity Matrix based on External Annotation}}
\label{sec:external_oracle}
Constructing $\mathcal{S}_F$ is straightforward when external pairwise judgments are available on whether two individuals $i, j$ should be treated similarly given a specific task \cite{lahoti2019operationalizing}.
The entry $S^F_{ij}$ in $\mathcal{S}_F$ is $1$ when individual $i$ and $j$ are labeled as similar, 
and $0$ otherwise. In this case, $\mathcal{S}_F$ is the adjacency matrix for the fairness graph $\mathcal{G}_F$. 
An alternative type of judgments is to map individuals into binary equivalence classes. A pair of individuals $i, j$ is linked in the fairness graph, $\mathcal{G}_F$, only if they belong to the same class \cite{lahoti2019operationalizing}. 

\paragraph{\textbf{Oracle Similarity Matrix based on Input Features}}
Although the similarity for individual fairness was originally envisioned to be provided by human experts \cite{dwork2012fairness}, it is often impractical to obtain for real-world tasks.
Previous works \cite{dong2021individual, 10.5555/3524938.3525596} obtain the oracle similarity matrix $\mathcal{S}_F$ from input feature space
i.e., the entry $S^F_{ij}$ in $\mathcal{S}_F$ measures the feature similarity between nodes $i$ and $j$, such as cosine similarity.
To construct the fairness graph, $\mathcal{G}_F$, we can discretize the continuous cosine values by connecting each node only to its top-$K$ similar nodes from 
$\mathcal{S}_F$.

\subsection{Fairness Hint Learning}\label{sec:fair-rep}
To incorporate the fairness information 
into the original GNN model, we learn fairness node representation with a separate GNN model using an auxiliary link prediction task on the fairness graph, $\mathcal{G}_F$. Similarly to masked language modeling in NLP to learn the contextual word embeddings by randomly masking words \cite{devlin2018bert}, we expect that predicting whether two nodes share a ``similarity'' edge inside the fairness graph can yield a valid fairness hint. We provide a theoretical analysis of this expectation later.

We use Graph Convolutional Network (GCN) model \cite{kipf2016semi} with GCN layers for link prediction, i.e., predicting whether two nodes in $\mathcal{G}_F$ share an edge. 
The initial input node features of the link prediction model are the same as the features of the original task. 
For any node embedding $h_{i}^l$ (the embedding of node $i$ from the $l$th hidden layer), GCN layer combines the node embedding $h_{i}^l$ and other node embeddings from its neighbor node set $\mathcal{N}(i)$, which is formally denoted as
\begin{equation}
\label{eq:gcn}
    h_{i}^{l + 1} = GCN(h_{i}^{l}, \{h_{j}^{l}, j\in \mathcal{N}(i)\})
\end{equation}
If setting the output of the last layer in the link prediction model for node $i$ and node $j$ as $v_{i}^f$ and $v_{j}^f$ respectively, we use the inner product and sigmoid function $\sigma(v_{i}^f \cdot v_{j}^f)$ as the probability of node $i$ and node $j$ sharing an edge, as well as the similarity between the outcome. We use cross entropy loss to optimize the link prediction model. 

We train this fairness representation learning model separately from the original task GNN model to avoid overfitting. We extract the output $v_{i}^f$ of the last layer for each node as the fairness hint, and the fairness hint is reusable for different backbone GNN models.
As we show next,
after training a link prediction model on the fairness graph, we can obtain individually fair fairness hints. When we feed the fairness hints to the original GNNs, the models can receive similar fairness hints for similar individuals and render similar outcomes. 


\textbf{Theoretical Analysis}
In our link prediction model, we use the inner product and sigmoid function as the similarity measure between the outcome and the cross entropy loss to optimize the GCN model, so the corresponding loss can be calculated as
\begin{equation}
\begin{split}
    & P_{ij} =  \sigma(v_{i}^f \cdot v_{j}^f)\\
    & \mathcal{L}_{link}(i, j) = y_{ij} \log (P_{ij}) + (1 - y_{ij}) \log(1 - P_{ij})
\end{split}
\end{equation}
where $y_{ij}$ is the label of the fairness edge existence and $P_ij$ is the probability of link existence between node $i$ and node $j$. Ideally, after optimizing the GNN model to minimize the cross entropy loss to 0, $P_{ij}$ should be 1 for two nodes connected by an edge and 0 for two nodes without any edge. We can then make an assumption that after the model training converges,  $P_{ij} < \delta$ if $y_{ij} = 0$ and $P_{ij} > 1 - \delta$ if $y_{ij} = 1$, for any $i$ and $j$ where $\delta$ is a small number. Given the inner product as the similarity measure, we can define the distance function of model outputs as $D(v_{i}^f, v_{j}^f) = 1 -\sigma(v_{i}^f \cdot v_{j}^f)$. The following conditions should hold.
\begin{equation}
    \begin{split}
        0 < D(v_{i}^f, v_{j}^f) < \delta, \quad&\textit{if}\quad y_{ij} = 1 \\
        1 - \delta < D(v_{i}^f, v_{j}^f) < 1, \quad&\textit{if}\quad  y_{ij} = 0
    \end{split}
\label{equ:assumption}
\end{equation}
With Definition \ref{def:fair} and Equation \ref{equ:assumption}, we can deduce the following theorem.
\begin{theorem}
Let $i$ and $j$ be two arbitrary nodes in a graph with node features $x_i$ and $x_j$. With Definition \ref{def:fair}, denote $\epsilon$ as the fairness tolerance. Assume that a GNN model for link prediction on this graph is trained with the similarity function $\sigma(v_{i}^f \cdot v_{j}^f)$ and then define the output distance function $D(v_{i}^f, v_{j}^f) = 1 -\sigma(v_{i}^f \cdot v_{j}^f)$. After training, we assume that $D(v_{i}^f, v_{j}^f) < \delta$ if an edge exists between $x_i$ and $x_j$ and $1 - \delta < D(v_{i}^f, v_{j}^f) < 1$ if no edge, where $\delta > 0$ is a small number. The outputs $v_{i}^f$ and $v_{j}^f$ are \textbf{individually fair} if $\epsilon > \delta$.
\label{theorem}
\end{theorem}
We provide the proof and the justification of how the assumption $\epsilon > \delta$ is achievable in practice in Appendix \ref{sec:proof}. The intuition behind this theorem is that, with the current representation learning setup, if nodes $i$ and $j$ share an edge in $\mathcal{G}_F$, the generated embeddings $v_{i}^f$ and $v_{j}^f$ from the link prediction model should be more similar than the other nodes without edges and thus similar nodes have similar fairness hints as the representations.
Theorem \ref{theorem} supports that our representation learning method encodes the fairness information of each node with the fairness hint. Next, we describe how to integrate the fairness hints with the learning of backbone GNN model for node classification tasks.

\subsection{Fairness Promotion for GNN Models}
\label{sec:fairness_promotion}

Our \textbf{GFairHint} framework is compatible with various GNN model architectures for the original tasks. The basic operations of each GNN layer are similar to the GCN operation in Equation (\ref{eq:gcn}), but the convolutional operations are replaced with other message-passing mechanisms for different GNN models. 
We train the chosen GNN backbone models with utility loss $\mathcal{L}_{utility}$ (i.e., cross entropy loss) and obtain the \textbf{utility node embedding} $u_{i}$ from the last GNN hidden layer. We then concatenate $u_{i}$ with the \textbf{fairness hint} $v_{i}^f$ to form a joint node embedding $[u_{i}, v_{i}^f]$. We add two layers of multilayer perceptron (MLP) with weights $W_1$ and $W_2$ to encourage the model to learn both utility and fairness information from the joint node embeddings. The final embedding $z_i$ of the node $i$ can be calculated as
\begin{equation*}
    z_i = W_2(W_{1}[u_{i}, v_{i}^f] + b_1) + b_2
\end{equation*}
We extract the fairness hint $v_{i}^f$ using the learned fairness representation learning model before training the original GNN model, and the fairness hint is fixed during the optimization process.
For node classification tasks, we apply \textit{softmax} to the final node embedding $z_i \in \mathbb{R}^c$ to obtain the predictions where $c$ is the number of classes and apply $\mathcal{L}_{utility}$ to optimize the parameters in the backbone GNN models. 
We empirically show that the fairness hint is fully involved in models' decision-making process via gradient-based interpretability method in Section \ref{sec:gradient_analysis} and the MLP layers can learn to balance between utility and fairness hint when we integrate the loss function with fairness regularization which we introduce next.

\subsection{Extension: Integration with Fairness Loss}
\label{sec:integration}
Our GFairHint framework can simply utilize the utility loss $\mathcal{L}_{utility}$ as the final loss.
Moreover, we can further encourage the model to learn fairness information by adding the ``fairness'' loss into the final loss function. Previous fairness promotion methods have designed various fairness loss to enforce a good balance between utility and fairness \cite{dong2021individual, NEURIPS2021_d9fea4ca}, which are complementary to our fairness hint.
In our work, we integrate the ranking-based fairness loss in REDRESS \cite{dong2021individual} to our framework GFairHint.

The objective of the loss is to minimize the difference between the oracle similarity matrix $\mathcal{S}_F$ and the outcome similarity matrix $\mathcal{S}_{\hat{Y}}$.
For each node, when $\mathcal{S}_F$ is based on input features, we can obtain two top-k ranking lists derived from $\mathcal{S}_F$ and $\mathcal{S}_{\hat{Y}}$ respectively. The fairness loss of node $i$ can be calculated as
\begin{align}
    & \hat{P}_{j, m}(i) = \frac{1}{1 - e^{(\hat{s}_{i, j} - \hat{s}_{i, m})}} \\
    & P_{j, m}(i) = \begin{cases}
          1 \quad &\text{if} \, s_{i, j} > s_{i, m} \\
          0.5 \quad &\text{if} \, s_{i, j} = s_{i, m} \\
          0 \quad &\text{if} \, s_{i, j} < s_{i, m} \\
     \end{cases}\\
    & \mathcal{L}_{j, m}(i) = -P_{j,m} \log \hat{P}_{j, m} - (1-P_{j,m})(1-\log \hat{P}_{j, m})\\
    & \mathcal{L}_{fairness}(i) = \sum_{j, m} \mathcal{L}_{j, m}(i) |\Delta z_{@k}|_{j, m}
\end{align}
where $z_{@k}(\cdot, \cdot)$ is the similarity metric (NDCG or ERR \cite{chapelle2009expected, jarvelin2002cumulated}) between the two top-k ranking lists and node $j$ or $m$ are selected from the ranking lists. Specifically, $s_{i, j}$ and $\hat{s}_{i, j}$ are the entries from $\mathcal{S}_F$ and $\mathcal{S}_{\hat{Y}}$ respectively.
However, when $\mathcal{S}_F$ is from external annotations where only pairwise judgements are available, the entries $S^F_{ij}$ in $\mathcal{S}_F$ are binary. Therefore, it is meaningless to calculate the ranking-based loss of the constructed fairness graph $\mathcal{G}_F$ in this case. In this work, to apply REDRESS related models to fairness datasets with external annotations, we made an adjustment by replacing $\mathcal{S}_F$ from external annotations with the one derived from input features. Details will be discussed in Section \ref{sec:baseline}.

The total fairness loss $\mathcal{L}_{fairness}$ is then calculated as the sum of the fairness loss on each node. The final objective is to combine the fairness loss and the utility loss.
\begin{equation}
\label{eq:loss}
    \mathcal{L}_{total} = \mathcal{L}_{utility} + \gamma \mathcal{L}_{fairness}
\end{equation}
where $\gamma$ is an adjustable hyperparameter. By changing the value of $\gamma$, we can control the weight of fairness and utility during training according to the task requirement. The details discussion is in Section \ref{sec:tradeoff}.



\section{Experiment Setup}
\subsection{Dataset Collection}
In our work, we focus on the node classification task to evaluate the fairness promotion ability of our proposed GFairHint. 
We collect five real-word datasets to assess the model performance in multiple domains (see statistics in Table \ref{tab:dataset}).
Coauthor-CS (CS) and Coauthor-Phy (Phy) are two co-authorship network datasets \cite{shchur2018pitfalls}, where each node represents an author, and they connect the nodes if two authors have published a paper together. 
ACM is a dataset of citation network \cite{tang2008arnetminer}, where each node represents a paper, and the edge denotes the citation relationship. 
These three datasets (ACM, CS, Phy) are also applied in the REDRESS paper as the experiment benchmarks \cite{dong2021individual}. In addition, we use another citation network OGBN-ArXiv dataset \cite{hu2020open}, which is several magnitude orders larger than the ACM, CS, and Phy datasets. 

For ACM, CS, and Phy datasets, we follow the preprocessing procedure in REDRESS and use the bag-of-word model to transform the title and abstract of a paper as its node feature. We use the pre-split training, validation, and test datasets from the REDRESS paper\footnote{\url{https://github.com/yushundong/REDRESS/tree/main/node\%20classification/data}}.
Regarding the ArXiv dataset, we directly use the processed 128-dimensional feature vectors from a pre-trained skip-gram model \cite{mikolov2013distributed}. We then follow the train/validation/test splits from the official release of Open Graph Benchmark (OGB).\footnote{\url{https://ogb.stanford.edu/docs/nodeprop/}} We repeat the experiments for each model setting twice, because the split of the dataset is fixed by the previous work. Since the citation and co-authorship network datasets do not contained human annotated similarity, we follow previous work \cite{dong2021individual} and use the cosine similarities between node features as the entries in $\mathcal{S}_F$. \footnote{We also tested with euclidean distance. We do not develop or evaluate different individual fairness measures but rather show that our method is compatible with various existing measures.}

Additionally, we curate a dataset with external human annotation on individual fairness similarity in the binary setting to demonstrate our framework's compatibility when external annotation is available.
The Crime dataset \cite{misc_communities_and_crime_183} consists of socioeconomic, demographic and law / police data records for neighborhoods in the US. We follow \citet{lahoti2019operationalizing} for most of the preprocessing and introduce additional information on the geometric adjacency of the county\footnote{https://pypi.org/project/county-adjacency/} to form a graph-structured dataset. The nodes are the neighborhoods, and the edges indicate that two neighborhoods reside in the same county or adjacent counties.
We have a binary outcome variable for whether the neighborhood is violent and consider other data records as input features.
For the similarity measure of individual fairness, we also follow \cite{lahoti2019operationalizing} to collect human reviews on Crime $\&$ Safety for neighborhoods in the U.S. from a neighborhood review website, Niche.\footnote{http://niche.com} 
The judgments are given in the form of 1-star to 5-star ratings by current and past residents of these neighborhoods. 
We then use aggregated mean ratings to construct the fairness graph as described in Section \ref{sec:external_oracle}, where the neighborhoods with the same rating level (e.g., 5 stars) are linked in the fairness graph.
As there is no predefined train/validation/test split, we randomly split the dataset and repeat five times for each model setting.

\begin{table}[t]
\centering
\caption{Statistics of the datasets used for node classification experiments. \# Nodes stands for the number of nodes for training dataset. 
}
\vspace{-1em}
\label{tab:dataset}
\resizebox{\linewidth}{!}{%
\begin{tabular}{ccccc} 
\toprule
$\mathcal{S}_F$ Type & Dataset & \# Nodes & \# Features & \# Classes \\ 
\midrule
\multirow{4}{*}{\begin{tabular}[c]{@{}l@{}}Input \\Feature\end{tabular}} & Coauthor-CS & 916 & 6,805 & 15 \\
 & Coauthor-Phy & 1,724 & 8,415 & 5 \\
 & ACM & 824 & 8,337 & 9 \\
 & ArXiv & 90,941 & 128 & 40 \\ 
\cmidrule{1-5}
External & Crime & 1,994 & 122 & 2 \\
\bottomrule
\end{tabular}
}
\end{table}




\begin{table*}[t]
\caption{Node classification results for citation datasets: ArXiv and ACM with cosine similarity as similarity measures. The number of layers and the hidden layer dimension of the backbone GNN models are 10 and 128 respectively. All values are reported in percentage. The first-best performance is marked in bold, and the second-best performance is underlined.}
\begin{tabular}{lllccc}
\hline
\textbf{Dataset}        & \textbf{Backbone}     & \textbf{Method}     & \textbf{Fairness: NDCG@10} & \textbf{Fairness: ERR@10} & \textbf{Utility: ACC} \\ \hline
\multirow{18}{*}{ArXiv} & \multirow{6}{*}{GCN}  & Vanilla             & 75.01 ± 0.19               & 91.45 ± 0.01              & {\ul 70.19 ± 0.02}    \\
                        &                       & InFoRM              & 74.66 ± 0.05               & 91.47 ± 0.09              & 68.86 ± 0.68          \\
                        &                       & REDRESS             & 75.25 ± 0.71               & 91.57 ± 0.10              & 68.65 ± 1.13          \\
                        &                       & REDRESS + MLP       & 75.01 ± 0.26               & 91.47 ± 0.01              & 69.85 ± 0.27          \\
                        &                       & GFairHint           & {\ul 81.93 ± 0.27}         & {\ul 94.28 ± 0.09}        & \textbf{70.62 ± 0.91} \\
                        &                       & GFairHint + REDRESS & \textbf{85.48 ± 3.47}      & \textbf{95.22 ± 0.80}     & 69.80 ± 0.47          \\ \cline{2-6} 
                        & \multirow{6}{*}{SAGE} & Vanilla             & 75.47 ± 0.37               & 91.71    ± 0.13           & \textbf{70.44 ± 0.69} \\
                        &                       & InFoRM              & 74.82 ± 0.18               & 91.60 ± 0.14              & 69.20 ± 1.45          \\
                        &                       & REDRESS             & 75.51 ± 0.73               & 91.58 ± 0.16              & 69.34 ± 0.55          \\
                        &                       & REDRESS + MLP       & 74.45    ± 0.53            & 91.40 ± 0.09              & 69.75 ± 0.18          \\
                        &                       & GFairHint           & {\ul 81.92 ± 0.17}         & {\ul 94.34 ± 0.04}        & {\ul 70.40 ± 0.34}    \\
                        &                       & GFairHint + REDRESS & \textbf{85.32 ± 3.45}      & \textbf{95.22 ± 0.79}     & 68.98 ± 0.25          \\ \cline{2-6} 
                        & \multirow{6}{*}{GAT}  & Vanilla             & 76.64 ± 0.21               & 92.04 ± 0.09              & {\ul 70.86 ± 0.64 }         \\
                        &                       & InFoRM              & 76.37 ± 0.04               & 91.92 ± 0.11              & 69.73 ± 0.40          \\
                        &                       & REDRESS             & 77.46 ± 0.09               & 92.18 ± 0.02              & 69.74 ± 0.19          \\
                        &                       & REDRESS + MLP       & 76.23 ± 0.98               & 91.86 ± 0.25              & 70.45 ± 0.30          \\
                        &                       & GFairHint           & {\ul 81.80 ± 0.14}         & {\ul 94.20 ± 0.04}        & \textbf{71.06 ± 0.45} \\
                        &                       & GFairHint + REDRESS & \textbf{85.49 ± 4.73}      & \textbf{95.20 ± 1.19}     & 69.89 ± 0.11    \\ \hline
\multirow{15}{*}{ACM}   & \multirow{5}{*}{GCN}  & Vanilla             & 33.90 ± 0.73               & {\ul 76.99 ± 0.08}        & \textbf{70.78 ± 0.18} \\
                        &                       & REDRESS             & 34.82 ± 0.80               & 76.98 ± 0.13              & 70.15 ± 1.77          \\
                        &                       & REDRESS + MLP       & 30.93 ± 0.46               & 76.66 ± 0.19              & {\ul 70.64 ± 1.89}    \\
                        &                       & GFairHint           & {\ul 35.12 ± 0.34}         & 76.39 ± 0.52              & 69.70 ± 0.77          \\
                        &                       & GFairHint + REDRESS & \textbf{38.58 ± 2.85}      & \textbf{77.00 ± 0.16}     & 69.77 ± 0.95          \\ \cline{2-6} 
                        & \multirow{5}{*}{SAGE} & Vanilla             & 30.55 ± 1.86               & 76.63 ± 0.18              & {\ul 69.26 ± 0.60}    \\
                        &                       & REDRESS             & 31.58 ± 1.06               & 76.68 ± 0.04              & 68.23 ± 0.97          \\
                        &                       & REDRESS + MLP       & 28.73 ± 0.12               & 76.29 ± 0.78              & \textbf{69.32 ± 0.44} \\
                        &                       & GFairHint           & {\ul 36.12 ± 0.72}         & {\ul 76.39 ± 0.25}        & 69.24 ± 0.11          \\
                        &                       & GFairHint + REDRESS & \textbf{37.83 ± 3.78}      & \textbf{77.37 ± 0.55}     & 67.52 ± 0.16          \\ \cline{2-6} 
                        & \multirow{5}{*}{GAT}  & Vanilla             & 34.62 ± 0.28               & 77.00 ± 0.20              & \textbf{71.14 ± 1.14} \\
                        &                       & REDRESS             & 34.83 ± 0.45               & {\ul 77.40 ± 0.28}        & 70.49 ± 0.87          \\
                        &                       & REDRESS + MLP       & 32.82 ± 1.36               & 76.22 ± 0.09              & 69.87 ± 0.70          \\
                        &                       & GFairHint           & {\ul 37.52 ± 0.54}         & 76.79 ± 0.27              & {\ul 71.04 ± 0.74}    \\
                        &                       & GFairHint + REDRESS & \textbf{43.01 ± 2.02}      & \textbf{77.50 ± 0.36}     & 69.65 ± 0.88          \\ \hline 
\end{tabular}%
\label{tab:result_acm_ogbn}
\end{table*}

\subsection{Methods for Comparison}
\label{sec:baseline}
To show the superiority of our proposed framework, we implement the vanilla GNN models and previous SOTA as baseline models with sensitivity analysis. 
Note that some existing works \cite{bose2019compositional, rahman2019fairwalk} for group fairness promotion cannot be used as baseline models because our work focuses on individual fairness. 
We explain these \textbf{baseline methods} below.

\textbf{Vanilla}: Vanilla denotes the vanilla GNN models without any individual fairness promotion method. 

\textbf{PFR}: PFR \cite{lahoti2019operationalizing} learns fair representation and can be considered as a pre-processing method for GNN models, and we adapt the implementation of PFR\footnote{https://github.com/plahoti-lgtm/PairwiseFairRepresentations} to GNN models by transforming the input features to the fairness representations and use the transformed representations as node features in vanilla GNN models. The PFR method requires computing the Laplacian matrix and the eigenvectors of the oracle similarity matrix $\mathcal{S}_F$ \cite{lahoti2019operationalizing}, so we have to explicitly store $\mathcal{S}_F$ in memory. The experiment of PFR on the Arxiv dataset with 90,941 nodes will cause out-of-memory issues. Since \citet{dong2021individual} has already shown REDRESS's priority over PFR on the ACM, Coauthor-Phy and Coauthor-CS datasets, we only experiment with PFR on the Crime dataset.

\textbf{InFoRM}: InFoRM \cite{kang2020inform} is a framework to promote individual fairness in conventional machine learning tasks on graphs. Similarly, due to the previous work's experiments showing the better performance of REDRESS than InFoRM on ACM, CS and Phy datasets, we only experiment with InFoRM on the Crime and Arxiv datasets by combining its fairness promotion loss with the utility loss of the GNN models.

\textbf{REDRESS}: REDRESS is the previous SOTA framework for individual fairness promotion in GNN models \cite{dong2021individual}. They formulate the conventional individual fairness promotion into a ranking-based optimization problem. By optimizing the ranking-based loss $\mathcal{L}_{fairness}$ and the utility loss $\mathcal{L}_{utility}$, REDRESS can achieve the goal of maximization of utility and promotion of individual fairness simultaneously. 
For the implementation of its framework and ranking-based loss, we adapt the codebase released by the authors\footnote{\url{https://github.com/yushundong/REDRESS}}.

\textbf{REDRESS + MLP} As mentioned in Section \ref{sec:fairness_promotion}, after concatenating the utility node embeddings and fairness hint, our proposed framework GFairHint uses additional MLP layers to process the concatenated embeddings, which increases the model complexity. This variant of REDRESS adds the MLP layers with the same size after the GNN models along with the original REDRESS loss for a fair comparison. We use the output of MLP layers from this variation model to calculate the loss and optimize the parameters in the GNN and MLP layers. REDRESS + MLP model can show the effectiveness of GFairHint without interference of the model complexity confounder.

\noindent\textbf{Our methods:} We study the performance of GFairHint and examine its effectiveness with the combination of REDRESS loss:

\textbf{GFairHint}: We combine the fairness hint with the utility node embedding and only use the $\mathcal{L}_{utility}$ loss to update the model parameters.

\textbf{GFairHint + REDRESS}: As described in Section \ref{sec:integration}, we combine the ranking-based loss $\mathcal{L}_{fairness}$ in REDRESS with the utility loss $\mathcal{L}_{utility}$ to further encourage the models to learn individual fairness. The only difference between this method and REDRESS + MLP is that GFairHint + REDRESS incorporates the fairness hint. 

We note that for the Crime dataset, since the entries of oracle similarity matrix $\mathcal{S}_F$ are binary (0-1), we cannot calculate the ranking-based loss of the constructed fairness graph $\mathcal{G}_F$. Therefore, we adapt the REDRESS-related methods to calculate the ranking-based loss based on input feature similarity. As a result, for the Crime dataset, REDRESS and REDRESS + MLP do not have any access to the fairness information (i.e., fairness graph $\mathcal{G}_F$), while GFairHint + REDRESS get fairness information only through the fairness hint but not the ranking-based loss.


\subsection{Evaluation Metric}
We evaluate both the \emph{utility performance} and the \emph{fairness performance} of the models. A desired model should achieve the highest fairness performance without sacrificing much, if it must, utility performance. For utility performance, we follow previous work \cite{dong2021individual} and the official OGB leaderboard to use classification accuracy (ACC). As for fairness performance, we use different evaluation metrics in accordance with two different settings of the oracle similarity, i.e., continuous and binary. 

For the co-authorship and citation networks (ACM, ArXiv, CS, Phy), the oracle similarity matrix is continuous, where the entry is the input feature similarity.
We follow previous work \cite{dong2021individual} to utilize ERR@K \cite{chapelle2009expected} and NDCG@K \cite{jarvelin2002cumulated}, where $k$ is the same as the threshold of to determine the edge existence in fairness graph in Section \ref{sec:fair-graph} for $\mathcal{S}_F$ based on input features. Higher ERR and NDCG values represent better individual fairness promotion.
These two metrics measure the similarity between the ranking lists obtained from the oracle similarity matrix $\mathcal{S}_F$ and the outcome similarity matrix $\mathcal{S}_{\hat{Y}}$. 
In the following experiments, we choose $k = 10$ and report the results of NDCG@10 and ERR@10 for all models for comparison.
We also provide the sensitivity analysis of difference choices of $k$ values on the model performance in Appendix \ref{sec:k_values}.

For the Crime dataset, where the entry of the oracle similarity matrix is binary, we follow \cite{lahoti2019operationalizing} and use \textbf{Consistency} as the evaluation metric. 
It measures the consistency of outcomes between individuals who are similar to each other. 
The formal definition regarding a fairness similarity matrix $S_{F}$ is
\begin{equation*}
    Consistency = 1 - \frac{\Sigma_i \Sigma_j | y_i - \hat{y_j} | \cdot S^{F}_{ij}}{\Sigma_i \Sigma_j  S^{F}_{ij}} \qquad \forall i \neq j.
\end{equation*}
Consistency measures how the predictions align with the oracle fairness similarity $S_{F}$. ERR and NDCG measure a similar concept from a ranking perspective.

\subsection{Implementation Details}
\label{sec:implementation_details}
All the backbone GNN models and our auxiliary link prediction models are implemented in the Pytorch framework, especially the package PyTorch Geometric \cite{paszke2017automatic, Fey/Lenssen/2019}. 
For each of our five datasets, we experiment with two backbone GNN settings, small and large model size. For the small model size setting, the number of layers and the dimension of the embeddings in the hidden layers are set to 2 and 16. For the big model size setting, we set these two numbers to 10 and 128 respectively. 
For all experiments, we fix the values of the hyperparameters $\gamma$ and $k$ at 1 and 10 as suggested in the previous work \cite{dong2021individual}, where $\gamma$ is the weighting factor when integrating with the ranking-based loss (Equation \ref{eq:loss}) and $k$ is the number of top entries used to calculate the ranking loss and fairness evaluation metrics NDCG@K and ERR@K. 

Our learned fairness hint for each node contains individual fairness information, and we expect that the fairness hint helps promote fairness in various GNN models. We choose three popular GNN models: GCN \cite{kipf2016semi}, GraphSAGE \cite{hamilton2017inductive}, and Graph Attention Networks (GAT) \cite{velivckovic2017graph} to demonstrate the compatibility of GFairHint with various GNN model designs. Note that we do not need to relearn the fairness hint for the same dataset even if the backbone models have changed.
Other details of the implementation are in Appendix \ref{sec:training_epoch}.

\section{Experimental Results}
\label{sec:result}
We aim to answer the questions in the following sections. 
Section \ref{sec:gfairhint_results}: How well can GFairHint and its extension method promote individual fairness compared with other baselines with different similarity measures?
Section \ref{sec:gradient_analysis}: How important are the fairness hint for GNN models when making predictions?
Section \ref{sec:tradeoff}: How does the fairness constraint hyperparemter $\gamma$ influence the performance of GFairHint?
Section \ref{sec:efficiency}: How computationally efficient is GFairHint and how does it compare with other baselines?

\subsection{Effectiveness of GFairHint}
\label{sec:gfairhint_results}
In this section, we present the results of our proposed methods and baselines on collected datasets.
For each dataset, we choose model hyperparameter settings with better average utility between two large and small model size settings as described in Section \ref{sec:implementation_details}. 

\paragraph{\textbf{Oracle Similarity Matrix based on Input Feature}}
For the citation and coauthorship networks, we use the input feature similarity as the entry of the oracle similarity matrix to construct the fairness graph. 
The results are shown in Table \ref{tab:result_acm_ogbn} for the citation networks (i.e., Arxiv and ACM) and Table \ref{tab:result_cs_phy} Appendix \ref{sec:euclidean} for the co-authorship networks (i.e., Phy and CS). \footnote{We follow previous works \cite{dong2021individual, kang2020inform} to use cosine similarity as the oracle similarity measure. We additionally include the results and discussion of using similarity measure based on Euclidean distance in Appendix \ref{sec:euclidean}.} 
For utility performance, our proposed GFairHint and GFairHint + REDRESS models achieve comparable results with vanilla backbone GNN models and other fairness promotion models. Indeed, in 5 cases out of 6 experiments for co-authorship datasets, our models achieve the best utility performance. Regarding the fairness performance, our proposed models achieve the best fairness performance in nearly all settings, except for the ERR value of the GNN models on the Phy dataset. We also achieve comparable ERR values with the SoTA REDRESS model for the Phy dataset. Moreover, our proposed GFairHint also behave better than the REDRESS model in the most scenarios of the citation network dataset. InFoRM method does not imporve the fairness performance on the Arxiv dataset, which is consistent with the results on the other academic networks from the REDRESS paper \cite{dong2021individual}. 

\begin{table}[t]
\centering
\caption{Node classification results on the Crime dataset. Consistency measures the fairness of the model. The number of layers and the hidden layer dimension of backbone GNN models are 2 and 16 respectively. All values are reported in percentage. Best results are in bold, and second-best results are underlined.}
\vspace{-1em}
\resizebox{\linewidth}{!}{%
\begin{tabular}{llcc}
\hline
\textbf{Backbone}              & \textbf{Method}     & \textbf{Consistency}  & \textbf{Acc}          \\ \hline
\multirow{7}{*}{\textbf{GCN}}  & Vanilla             & 54.80 ± 0.23          & 73.83 ± 0.34          \\
                               & PFR                 & 52.20 ± 0.55          & 71.53 ± 1.10          \\
                               & InFoRM              & 56.84 ± 1.77          & 72.93 ± 0.96          \\
                               & REDRESS             & 54.07 ± 0.96          & 73.98 ± 0.70          \\
                               & REDRESS + MLP       & 53.06 ± 1.04          & 73.58 ± 1.80          \\
                               & GFairHint           & {\ul 62.76 ± 2.74}    & {\ul 75.44 ± 0.71}    \\
                               & GFairHint + REDRESS & \textbf{63.61 ± 4.44} & \textbf{75.54 ± 0.90} \\ \hline
\multirow{7}{*}{\textbf{SAGE}} & Vanilla             & 62.09 ± 0.50          & \textbf{82.16 ± 0.33} \\
                               & PFR                 & 56.75 ± 0.95          & 80.15 ± 0.67          \\
                               & InFoRM              & 60.93 ± 2.59          & 79.05 ± 0.51          \\
                               & REDRESS             & 61.46 ± 1.91          & {\ul 82.11 ± 0.52}    \\
                               & REDRESS + MLP       & 61.46 ± 1.36          & 81.35 ± 0.34          \\
                               & GFairHint           & {\ul 62.26 ± 0.98}    & 80.60 ± 0.98          \\
                               & GFairHint + REDRESS & \textbf{62.49 ± 4.86} & 80.85 ± 1.21          \\ \hline
\multirow{7}{*}{\textbf{GAT}}  & Vanilla             & 55.17 ± 0.81          & 73.68 ± 0.79          \\
                               & PFR                 & 54.06 ± 1.20          & 73.83 ± 0.90          \\
                               & InFoRM              & 53.44 ± 2.24          & 71.38 ± 1.43          \\
                               & REDRESS             & 53.55 ± 1.15          & 72.88 ± 0.74          \\
                               & REDRESS + MLP       & 51.84 ± 0.42          & 72.08 ± 1.24          \\
                               & GFairHint           & {\ul 64.04 ± 2.74}    & \textbf{75.34 ± 0.74} \\
                               & GFairHint + REDRESS & \textbf{65.30 ± 3.60} & {\ul 74.94 ± 1.05}    \\ \hline
\end{tabular}
}
\label{tab:crime}
\end{table}

\paragraph{\textbf{Oracle Similarity Matrix based on External Annotation}}
For the Crime dataset, we construct a fairness graph from collected human expert judgements. 
We show the results in Table \ref{tab:crime}. 
We find that
for all three backbone GNN models, GFairHint and GFairHint + REDRESS are the best two methods in the fairness (Consistency) evaluation. This is as expected since Vanilla and REDRESS models do not have access to fairness information in this setting, and InFoRM and PFR methods are not specifically designed for GNN models. 
GFairHint and GFairHint + REDRESS have close performance in consistency, demonstrating the effectiveness of fairness hint even when it is used alone. Although GFairHint + REDRESS has slightly better results, it has much higher computational cost because of the ranking-based loss. Detailed discussion on computation efficiency is in Section \ref{sec:efficiency}.
For GCN and GAT backbone models, our proposed methods achieve the best two results in utility (accuracy) evaluation.

\paragraph{\textbf{Summary}}
We systematically evaluate the utility performance and fairness performance in $5 \times 3 =15$ combinations of dataset and backbone model, which results in $15$ utility comparisons and $27$ fairness comparisons.\footnote{For the four academic networks with oracle similarity matrix based on input feature, we evaluated two fairness metrics, which leads to $(4 \times 2 + 1) \times 3 = 27$ comparisons for fairness}.
Our proposed GFairHint + REDRESS method achieved best fairness performance in almost all comparisons (24/27), while GFairHint performed second best in 16/27 of the comparisons when applied alone. 
These two methods also have comparable utility performance with the Vanilla model, as they ranked top two in 12/15 utility comparisons.
Although GFairHint + REDRESS achieved better fairness performance than GFairHint in general, the gaps are small. GFairHint even ranked higher in 6/15 utility performance, especially for large dataset (3/3).
These observations empirically show that GFairHint achieves a good balance between utility and fairness, and demonstrate that while GFairHint is better than previous work on individual fairness promotion in most cases, 
it is also complementary to other methods with fairness regularization loss and can further improve the performance.

\begin{table}[t]
\centering
\caption{\label{tab:gradient} Average importance scores of utility node embedding and fairness hint for each dataset with GAT backbone model. We also report the ratio of average importance scores of two types of node embeddings.}
\vspace{-1em}
\resizebox{\columnwidth}{!}{%
\begin{tabular}{llccc}
\hline
\textbf{Dataset} & \textbf{Method}              & \textbf{Score($v^f$)} & \textbf{Score($u$)} & \textbf{$\frac{\text{Score}(u)}{\text{Score}(v^f)}$} \\ \hline
\multirow{2}{*}{Arxiv} & GFairHint         & 0.044           & 0.041                   & 0.923                           \\
                       & GFairHint+REDRESS & 0.051           & 0.034                   & 0.677                           \\ \hline
\multirow{2}{*}{ACM}   & GFairHint         & 0.158           & 0.056                   & 0.354                           \\
                       & GFairHint+REDRESS & 0.245           & 0.045                   & 0.184                           \\ \hline
\multirow{2}{*}{CS}    & GFairHint         & 0.157           & 0.164                   & 1.050                           \\
                       & GFairHint+REDRESS & 0.109           & 0.092                   & 0.848                           \\ \hline
\multirow{2}{*}{Phy}   & GFairHint         & 0.169           & 0.182                   & 1.073                           \\
                       & GFairHint+REDRESS & 0.225           & 0.166                   & 0.738                           \\ \hline
\multirow{2}{*}{Crime} & GFairHint         & 0.065           & 0.228                   & 3.536                           \\
                       & GFairHint+REDRESS & 0.022           & 0.057                   & 2.594                           \\ \hline
\end{tabular}
}
\end{table}

\subsection{Importance of Fairness Hint}
\label{sec:gradient_analysis}

In Section \ref{sec:fair-rep}, we prove that our learned fairness hints are individually fair and integrate the fairness hints into the training of backbone GNN models to achieve better fairness promotion. 
We further evaluate: \textit{Does GNN backbone model utilize the concatenated fairness hints in the node label prediction?}

We answer this question by borrowing the idea of saliency map 
\cite{simonyan2013deep} to demonstrate the relative importance of utility node embedding, $u$, and fairness hints, $v^f$, when making predictions.
We calculate the gradient of the model output w.r.t. each dimension of the node embedding as the importance score for the dimension. We then average the importance score for utility node embedding and fairness hint as $\text{Score}(u)$ and $\text{Score}(v^f)$ respectively. 
The details are shown in Appendix \ref{sec:hint_importance}.

From Table \ref{tab:gradient}, we observe that the magnitude of importance score of utility node embedding and fairness hint for our framework is comparable, which indicates that GNNs in our method actually apply the fairness hint to predict the node labels. 
Moreover, the values of $\frac{\text{Score}(u)}{\text{Score}(v^f)}$ for GFairHint + REDRESS are lower than the ones for GFairHint, suggesting that the fairness hint becomes more important when integrated with fairness loss. 
This further demonstrates that GFairHint is a plug-and-play framework as the fairness hint can be effectively utilized by the previous SoTA fairness promotion method with fairness regularization.

\subsection{Trade-off between Fairness and Utility}
\label{sec:tradeoff}
\begin{figure}[t]
\centering
    \includegraphics[width=\linewidth]{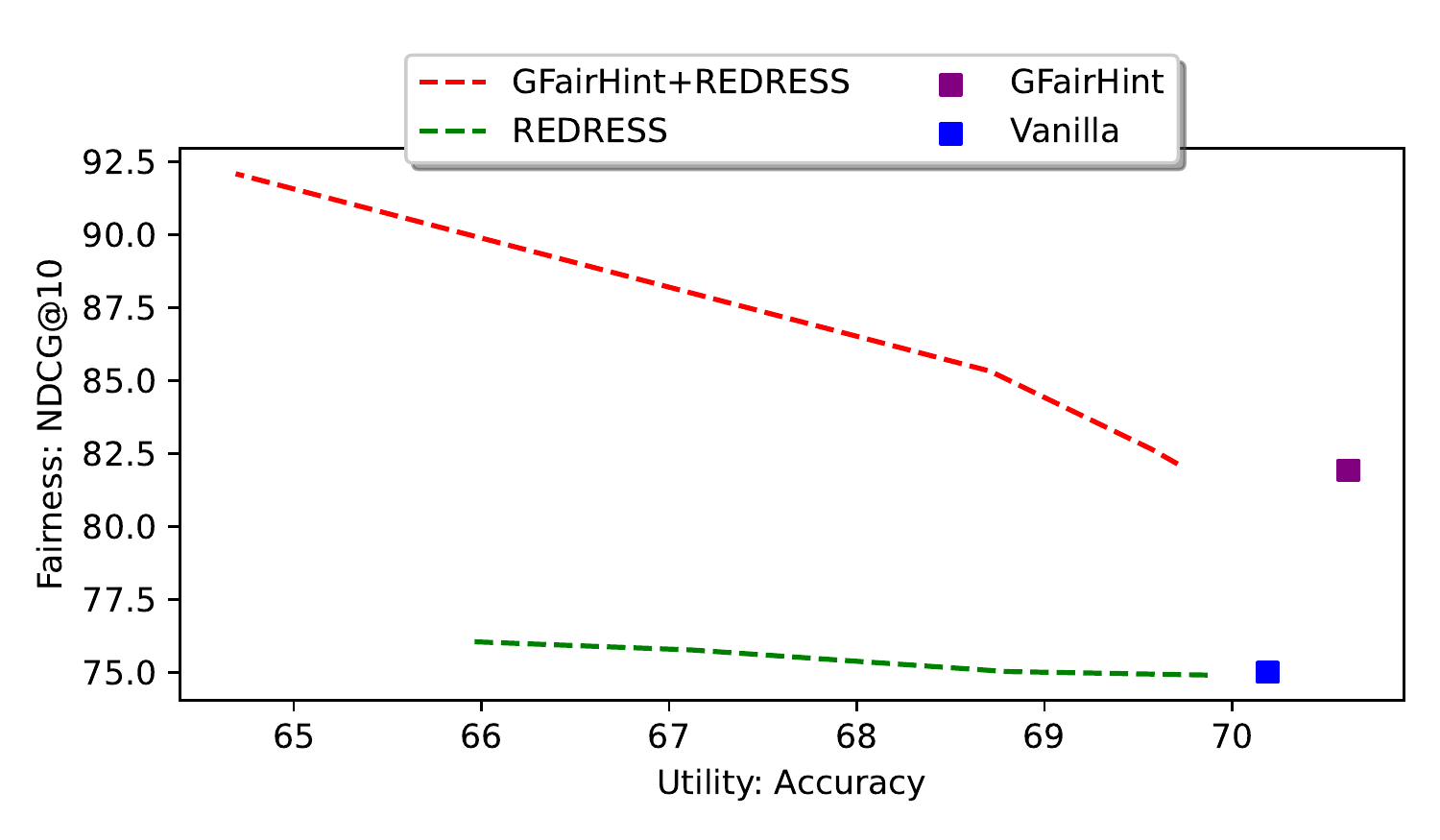}
    \vspace{-2em}
    \caption{Pareto frontiers of various methods on Arxiv dataset. Upper-right corner (high accuracy, high NDCG@10) is preferable. Our method outperforms baseline methods significantly in the trade-off between fairness and utility.}
    \label{fig:fairness_trade_off}
\end{figure}

GFairHint + REDRESS achieves the best fairness performance, where we integrate fairness hint with ranking-based loss. The value of the hyperparameter $\gamma$ in Equation \ref{eq:loss} controls the strength of the fairness constraint. There is a trade-off between utility and fairness when adjusting the $\gamma$ value \cite{dong2021individual}. 
To demonstrate the effectiveness of GFairHint, we perform experiments with multiple values of $\gamma$ for the REDRESS and GFairHint + REDRESS models on the Arxiv dataset with GCN as the backbone GNN model. 
Figure \ref{fig:fairness_trade_off} shows the trade-off between accuracy and fairness (NDCG@10) with varying values of $\gamma$ for the REDRESS and GFairHint + REDRESS methods. The curves in the figure for REDRESS and GFairHint + REDRESS are generated by changing a range of fairness coefficients $\gamma$ from 0.01 to 100 and computing the Pareto frontiers. We also further visualize the accuracy and NDCG@10 values for the vanilla and GFairHint models as two data points for reference.

With the value of $\gamma$ being small (e.g., $0.001$), REDRESS and GFairHint + REDRESS models behave similarly to the vanilla and GFairHint models respectively as expected. When increasing the value of $\gamma$, we can observe fairness improvements for both REDRESS and the GFairHint + REDRESS models. This fairness improvement is more significant for the GFairHint + REDRESS model. 
We conjecture that little improvement for the REDRESS model is due to the vanishing gradient problem of deep GNN models \cite{demir2021eeg}, which may reduce the impact of fairness loss. 
GFairHint + REDRESS model is less impacted because it also directly learns from the fairness hint that is incorporated into the final GNN layers. 
We observe that with the same accuracy level, our proposed GFairHint + REDRESS model achieves a higher NDCG@10 value than the REDRESS model, demonstrating the effectiveness of the fairness hint.

We expect that the adjustment of the trade-off between fairness and utility can provide more flexibility in practical applications. For example, some tasks may pay more attention to fairness rather than utility. 

\subsection{Efficiency Evaluation}
\label{sec:efficiency}

In addition to fairness and utility results, we compare the efficiency of GFairHint models with other baseline models in terms of time complexity and empirical training time. 
The time complexity of training a $l$-layer GCN model is $\mathcal{O}(l·n + l·|e|)$, where $|e|$ is the edge number and $n$ is the node number \cite{blakely2021time}. During the training of GNN model, the ranking-based loss in REDRESS requires finding a list containing top-k similar nodes for each node and ranking the list. The additional complexity of REDRESS is $\mathcal{O}(n\cdot\log(n)\cdot k)$ \cite{dong2021individual} and our additional time complexity is $\mathcal{O}(n)$ since we only add two MLP layers when training GNN models

The additional cost introduced by the auxiliary link prediction task is small because the complexity is the same as the original node classification task and the learned fairness can be reused for other backbone GNN models on the same dataset, which reduces the marginal cost of this auxiliary task.
Moreover, even if the fairness hint is only used once, the time required for training a link prediction GNN and a node classification GNN is significantly smaller than using the REDRESS framework. We empirically show the difference in training time in Figure \ref{fig:training_time} and the details in Appendix \ref{sec:efficiency_supp}. The results show that the additional cost of GFairHint is negligible compared to the Vanilla model. Therefore, the proposed GFairHint model is more scalable in practice when applied to large graph datasets. 



\section{Limitations and Future Work}
One crucial question for individual fairness is the source of similarity measures. 
External annotation is often impractical, subjective, and potentially biased, and the input feature can be an incomplete and imperfect source.
It requires comprehensive domain knowledge to develop the fairness similarity measure for a specific real-world application.
We note that our framework is compatible with different types of similarity measures as long as the construction of a fairness graph is viable. 
Moreover, there may be bias in the original input features, and we can have a sense of such bias by looking at the fairness evaluations of the vanilla models. 
However, it is unavoidable to use original information, as it is the only source of feature information available for making predictions. The intuition for promoting individual fairness is to use oracle individual fairness similarity to guide the model to utilize such fairness information, whereas in this paper we rely on the learned fairness hint. We acknowledge that there are works that debias the original information directly as pre-processing \cite{kang2020inform, lahoti2019operationalizing}, but they are not as empirically effective as the proposed method or the in-processing method \cite{dong2021individual} when applied along. Our method is also orthogonal to these pre-processing methods and is thus possible to integrate with them for better performance. 
Lastly, in this paper, we show the effectiveness of fairness hint with a simple concatenation strategy, and it is possible to develop more complex and better integration methods to utilize fairness hint, especially in compliance with the tasks and data formats at hand.
We leave these directions for future work.

\section{Conclusions}
In this work, we propose \textbf{GFairHint}, a plug-and-play framework for promoting individual fairness in GNNs via fairness hint. 
Our method learns fairness hint through an auxiliary link prediction task on a constructed fairness graph. 
The fairness graph can be derived from both continuous and binary oracle similarity matrix, corresponding to two ways of obtaining similarity for individual fairness respectively, i.e., from input feature space and from external human annotations. 
We also integrate GFairHint with another complementary individual fairness promotion method, REDRESS.
We conduct extensive empirical evaluations on node classification tasks to show the effectiveness of our proposed method in achieving good balance in utility and fairness, with much less computational cost.


    
    
    
    
    

\bibliographystyle{ACM-Reference-Format}
\bibliography{sample-base}

\appendix

\section{Training Process Details}
\label{sec:training_epoch}
When training the GCN model for the auxiliary link prediction task, we use a two-layer GCN model with embedding size as 128. We randomly mask the $2.5\%$ and $5\%$ edges of the fairness graph as the positive edges sampled in the validation set and the test set. We also generate the same number of negative edges. For optimization, we use Adam optimizer \cite{kingma2014adam} with learning rate 0.001 and full batch training for 200 epochs. 

For the main node classification task, when training without the ranking-based loss (Vanilla and GFairHint methods), the numbers of training epochs of ArXiv, ACM, Phy, CS and Crime datasets are 300, 150, 300, 300, 500 respectively. When training with ranking-based loss, we first train the models with only utility loss for tens of epochs to ``warm up" and then the models will be trained with ranking-based loss and utility loss together. The numbers of ``warm-up" epochs and training epochs with ranking-based loss are 150 and 300, 250 and 300, 50 and 150, 50 and 500, 50 and 600 for ArXiv, ACM, Phy, CS and Crime datasets, respectively. This warm-up operation also follows the procedure in the REDRESS paper \cite{dong2021individual}.

The hyperparameters for model size are as follows. 
For citation networks, the results are from 10-layer models with hidden layer dimension 128. For co-authorship netowrks, the results are from 2-layer models with hidden layer dimension 16.
Specifically, for the ArXiv dataset, since they have about 90k nodes in the training dataset, using 10-layer GAT would cause a memory issue, so we only experiment on the Arxiv data with the 3-layer and 128-dimensional hidden layer GAT model.
For Crime network, we select the small model size setting with 2 hidden layers and hidden layer dimension size as 16.

\begin{table*}[h]
\caption{Node classification results on co-authorship datasets: Coauthor-Phy and Coauthor-CS with cosine similarity as similarity measures. The number of layers and the hidden layer dimension of backbone GNN models are 2 and 16 respectively. All values are reported in percentage. The first-best performance is marked in bold, and the second-best performance underlined.}
\label{tab:result_cs_phy}
\begin{tabular}{lllccc}
\hline
\textbf{Dataset}      & \textbf{Backbone}     & \textbf{Method}     & \textbf{Fairness: NDCG@10} & \textbf{Fairness: ERR@10} & \textbf{Utility: ACC} \\ \hline
\multirow{15}{*}{CS}  & \multirow{5}{*}{GCN}  & Vanilla             & 44.00 ± 1.14               & 78.93 ± 0.07              & 80.16 ± 9.32          \\
                      &                       & REDRESS             & 49.24 ± 2.36               & {\ul 81.25 ± 0.55}        & 79.88 ± 2.68          \\
                      &                       & REDRESS + MLP       & 40.54 ± 2.45               & 78.76 ± 0.27              & 77.35 ± 2.10          \\
                      &                       & GFairHint           & {\ul 51.31 ± 1.17}         & 79.56 ± 0.36              & {\ul 87.08 ± 2.05}    \\
                      &                       & GFairHint + REDRESS & \textbf{64.60 ± 0.58}      & \textbf{83.48 ± 0.20}     & \textbf{91.17 ± 0.54} \\ \cline{2-6} 
                      & \multirow{5}{*}{SAGE} & Vanilla             & 46.34 ± 0.98               & 78.33 ± 0.02              & 85.49 ± 6.58          \\
                      &                       & REDRESS             & {\ul 54.71 ± 1.91}         & {\ul 81.09 ± 0.59}        & {\ul 88.26 ± 3.30}    \\
                      &                       & REDRESS + MLP       & 42.83 ± 1.40               & 77.80 ± 0.44              & 83.29 ± 1.66          \\
                      &                       & GFairHint           & 51.00 ± 0.73               & 79.32 ± 0.30              & 86.67 ± 3.07          \\
                      &                       & GFairHint + REDRESS & \textbf{64.49 ± 0.45}      & \textbf{83.21 ± 0.31}     & \textbf{91.06 ± 0.02} \\ \cline{2-6} 
                      & \multirow{5}{*}{GAT}  & Vanilla             & 46.99 ± 0.98               & 79.44 ± 0.29              & 80.73 ± 7.52          \\
                      &                       & REDRESS             & 51.14 ± 0.25               & 80.62 ± 0.00              & 79.53 ± 2.75          \\
                      &                       & REDRESS + MLP       & 44.39 ± 0.73               & 79.13 ± 0.48              & 82.42 ± 1.87          \\
                      &                       & GFairHint           & {\ul 53.80 ± 0.99}         & {\ul 80.91 ± 0.38}        & {\ul 86.11 ± 0.94}    \\
                      &                       & GFairHint + REDRESS & \textbf{63.67 ± 0.09}      & \textbf{83.46 ± 0.33}     & \textbf{90.54 ± 0.57} \\ \hline
\multirow{15}{*}{Phy} & \multirow{5}{*}{GCN}  & Vanilla             & 30.46 ± 1.05               & 73.30 ± 0.10              & 88.33 ± 5.11          \\
                      &                       & REDRESS             & 35.76 ± 1.72               & \textbf{74.69 ± 0.06}     & 84.28 ± 2.12          \\
                      &                       & REDRESS + MLP       & {\ul 36.06 ± 0.88}         & {\ul 74.61 ± 0.13}        & {\ul 93.40 ± 0.38}    \\
                      &                       & GFairHint           & 33.26 ± 0.21               & 71.60 ± 0.32              & 87.35 ± 0.03          \\
                      &                       & GFairHint + REDRESS & \textbf{41.53 ± 4.63}      & 73.87 ± 1.46              & \textbf{94.15 ± 0.15} \\ \cline{2-6} 
                      & \multirow{5}{*}{SAGE} & Vanilla             & 30.66 ± 0.94               & 72.45 ± 0.15              & \textbf{94.88 ± 0.66} \\
                      &                       & REDRESS             & {\ul 41.01 ± 2.25}         & \textbf{74.89 ± 0.64}     & 89.72 ± 0.33          \\
                      &                       & REDRESS + MLP       & 35.65 ± 0.90               & {\ul 74.22 ± 0.04}        & 93.08 ± 0.04          \\
                      &                       & GFairHint           & 29.95 ± 0.98               & 71.75 ± 0.28              & 90.00 ± 3.15          \\
                      &                       & GFairHint + REDRESS & \textbf{41.66 ± 0.23}      & 74.06 ± 0.03              & {\ul 93.24 ± 0.69}    \\ \cline{2-6} 
                      & \multirow{5}{*}{GAT}  & Vanilla             & 33.27 ± 1.32               & 73.78 ± 0.57              & 90.33 ± 5.19          \\
                      &                       & REDRESS             & {\ul 36.24 ± 0.06}         & {\ul 74.64 ± 0.34}        & 84.74 ± 5.39          \\
                      &                       & REDRESS + MLP       & 36.43 ± 2.78               & \textbf{74.70 ± 0.97}     & {\ul 92.54 ± 0.15}    \\
                      &                       & GFairHint           & 30.16 ± 0.11               & 71.99 ± 0.45              & 88.10 ± 1.77          \\
                      &                       & GFairHint + REDRESS & \textbf{44.56 ± 2.62}      & 74.62 ± 1.10              & \textbf{93.67 ± 0.30} \\ \hline
\end{tabular}
\end{table*}

\begin{table*}[ht]
\caption{Node classification results on datasets: Arxiv, ACM, and Coauthor-Phy and Coauthor-CS with euclidean distance as similarity measures. The number of layers and the hidden layer dimension of backbone GNN models are the same as the setting in Table \ref{tab:result_acm_ogbn} and \ref{tab:result_cs_phy}. All values are reported in percentage. The first-best performance is marked in bold, and the second-best performance underlined.}
\label{tab:result_euclidean}
\begin{tabular}{llccc}
\hline
Dataset                & Method              & Fairness: NDCG@10     & Fairness: ERR@10      & Utility: ACC          \\ \hline
\multirow{5}{*}{Arxiv} & Vanilla             & 74.09 ± 0.49          & 48.85 ± 0.45          & {\ul 70.18 ± 0.04}    \\
                       & REDRESS             & 74.27 ± 0.42          & 49.15 ± 0.42          & 69.34 ± 0.77          \\
                       & REDRESS + MLP       & 72.71 ± 1.30          & 48.04 ± 1.17          & 69.08 ± 1.84          \\
                       & GFairHint           & {\ul 80.01 ± 0.46}    & {\ul 50.75 ± 0.48}    & \textbf{70.81 ± 0.47} \\
                       & GFairHint + REDRESS & \textbf{81.67 ± 1.59} & \textbf{51.82 ± 1.00} & 68.22 ± 2.61          \\ \hline
\multirow{5}{*}{ACM}   & Vanilla             & 37.37 ± 1.25          & 20.00 ± 0.46          & 69.28 ± 1.77          \\
                       & REDRESS             & 38.04 ± 1.44          & 20.24 ± 0.59          & \textbf{70.14 ± 0.92} \\
                       & REDRESS + MLP       & 34.51 ± 0.80          & 18.71 ± 0.38          & 68.86 ± 1.31          \\
                       & GFairHint           & {\ul 43.31 ± 0.39}    & {\ul 22.54 ± 0.19}    & {\ul 69.89 ± 1.09}    \\
                       & GFairHint + REDRESS & \textbf{43.94 ± 0.25} & \textbf{22.89 ± 0.13} & 69.63 ± 0.53          \\ \hline
\multirow{5}{*}{CS}    & Vanilla             & 42.94 ± 0.72          & 21.07 ± 0.31          & 83.43 ± 2.01          \\
                       & REDRESS             & 44.08 ± 0.50          & 21.57 ± 0.22          & \textbf{89.78 ± 0.86} \\
                       & REDRESS + MLP       & 27.99 ± 0.87          & 14.64 ± 0.42          & 72.85 ± 1.22          \\
                       & GFairHint           & {\ul 49.78 ± 1.42}    & {\ul 23.66 ± 0.58}    & 85.96 ± 0.45          \\
                       & GFairHint + REDRESS & \textbf{52.16 ± 0.29} & \textbf{24.45 ± 0.08} & {\ul 88.06 ± 0.75}    \\ \hline
\multirow{5}{*}{Phy}   & Vanilla             & 29.39 ± 0.14          & 17.68 ± 0.11          & {\ul 95.08 ± 0.20}    \\
                       & REDRESS             & {\ul 34.03 ± 0.66}    & {\ul 19.88 ± 0.33}    & \textbf{95.72 ± 0.06} \\
                       & REDRESS + MLP       & 24.56 ± 2.75          & 15.25 ± 1.51          & 87.69 ± 0.23          \\
                       & GFairHint           & 26.96 ± 1.50          & 16.38 ± 0.73          & 90.39 ± 1.09          \\
                       & GFairHint + REDRESS & \textbf{34.73 ± 0.72} & \textbf{20.15 ± 0.35} & 92.22 ± 1.46          \\ \hline
\end{tabular}
\end{table*}

\section{Supplementary Results}
\label{sec:euclidean}
To verify the robustness of our framework on the other measures of similarity based on input features, we reproduce the experiments for oracle similarity matrix based on input feature but replace the similarity measures from the cosine similarity to euclidean distance. The similarity $S_e(i, j)$ between two nodes $i$ and $j$ with by euclidean distance $D_e(i, j)$ can be calculated as follows:
\begin{equation}
    S_e(i, j) = 1 - \frac{D_e(i, j)}{\max\{D_e(i, j), \forall i, j \}}
\end{equation}
Here, we apply the SAGE model as the backbone GNN model and present the results of our framework and other baseline methods in Table \ref{tab:result_euclidean}. 

We can observe that the behaviour of our framework GFairHint and GFairHint + REDRESS is similar as its performance in Section \ref{sec:result}.  For utility performance, our proposed GFairHint and GFairHint + REDRESS models achieve comparable results with vanilla backbone GNN models and other fairness promotion models. Regarding the fairness performance, our proposed models GFairHint + REDRESS achieve the best fairness performance in all scenarios. Moreover, our proposed GFairHint also behaves better than the REDRESS model in the most scenarios except on the Coauthor-Phy dataset. Based on these observations, we can indicate that our findings in Section \ref{sec:result} can be generalized on other continuous similarity measures if the oracle similar matrix is based on the input features, which demonstrates the robustness of our framework on different types of similarity measures.

\section{Extended Theoretical Analysis}
\label{sec:proof}
In this section, we will first prove Theorem \ref{theorem} and provide the justification of the assumption $\epsilon > \delta$ is achievable in practice in Theorem \ref{theorem}.
\begin{proof}
Since $\epsilon$ in Definition \ref{def:fair} is a constant and can be set to an arbitrary number greater than 0, we can set $\epsilon$ as the edge existence threshold when constructing the fairness graph where edge exists only when the oracle similarity between two nodes is greater than $\epsilon$. 

We will separately discuss two scenarios for $x_i$ and $x_j$: with or without an edge between them on the fairness graph. For two nodes $x_i$ and $x_j$ on the fairness graph, if there is no edge between them, then we have $0 < S(x_i, x_j) \leq \epsilon$. The fairness bound and the outcome distance hold the condition that
\begin{equation}
       D(v_{i}^f, v_{j}^f) < 1 \leq \frac{\epsilon}{S(x_i, x_j)}
\end{equation}
The outcome of GNN models for $x_i$ and $x_j$ are individually fair.
Similarly, for two nodes $x_i$ and $x_j$, if there is an edge between them, then we have $\epsilon < S(x_i, x_j) < 1$. For the fairness bound and the outcome distance, we can deduce that
\begin{equation}
    D(v_{i}^f, v_{j}^f) < \delta < \epsilon < \frac{\epsilon}{S(x_i, x_j)} < 1 
\end{equation}
The individual fair condition still holds.
Hence, we can conclude that the outcomes (fairness hint) of our link prediction GNN model training on the fairness graph are individually fair.
\end{proof}

\paragraph{Justification of $\epsilon > \delta$} In both settings of oracle similarity matrix in Section \ref{sec:fair-graph}, only a small proportion of node pairs can be connected so the threshold of edge existence, $\epsilon$, is close to the maximum similarity score between any two nodes in the fairness graph. In practice, for datasets with oracle similarity from external annotation, $\epsilon = 1$, and for datasets with oracle similarity from input features, $\epsilon>0.9$. Meanwhile, the link prediction models for learning fairness hint achieved low loss (less than 0.1) in our experiments, which implies a low value of loss tolerance $\delta$. Hence, our assumption the fairness tolerance $\epsilon > \delta$ in Theorem \ref{theorem} should hold in most circumstances.

\section{Supplementary Material on Importance of Fairness Hint}
\label{sec:hint_importance}

Suppose that $S_k$ is the model output with respect to the input node $k$, and $[u_k, v_k^f]$ is the joint node embedding with fairness hint $v_k^f$ and utility node embedding $u_k$ of node $k$. $u_{ki}$ and $v_{ki}^f$ represents the $i$-th dimension of the utility node embedding and fairness hint respectively. The importance of the node embedding dimension $u_i$ can be viewed as:
\begin{equation}
    \text{Score}(u_{ki}) = \frac{\partial S_k}{\partial u_{ki}}
\end{equation}
Hence, we can define the importance of the utility node embedding and fairness hint as the average importance of each node embedding dimension,
\begin{equation}
        \text{Score}(u_k) = \frac{1}{n} \sum_{i=1}^n | \frac{\partial S_k}{\partial u_{ki}} | \quad \text{Score}(v_k^f) = \frac{1}{m} \sum_{i=1}^m | \frac{\partial S_k}{\partial v_{ki}^f} |
\end{equation}
where $n$ and $m$ is the dimension number of utility embedding and fairness hint respectively. For the whole dataset, we calculate the average embedding importance across all nodes and obtain the \textbf{overall utility or fairness importance}:
\begin{equation}
    \text{Score}(u) = \frac{1}{\lambda} \sum_{k=1}^\lambda \text{Score}(u_k) \quad \text{Score}(v^f) = \frac{1}{\lambda} \sum_{k=1}^\lambda \text{Score}(v_k^f)
\end{equation}
where $\lambda$ is the number of nodes in the dataset. We calculate the overall importance for utility node embedding and fairness hint for each dataset with respect to our GFairHint and GFairHint + REDRESS framework and the ratio between two importance scores.

\section{Supplementary Material on Efficiency Evaluation}
\label{sec:efficiency_supp}

\begin{figure}[t]
\centering
    \includegraphics[width=\linewidth]{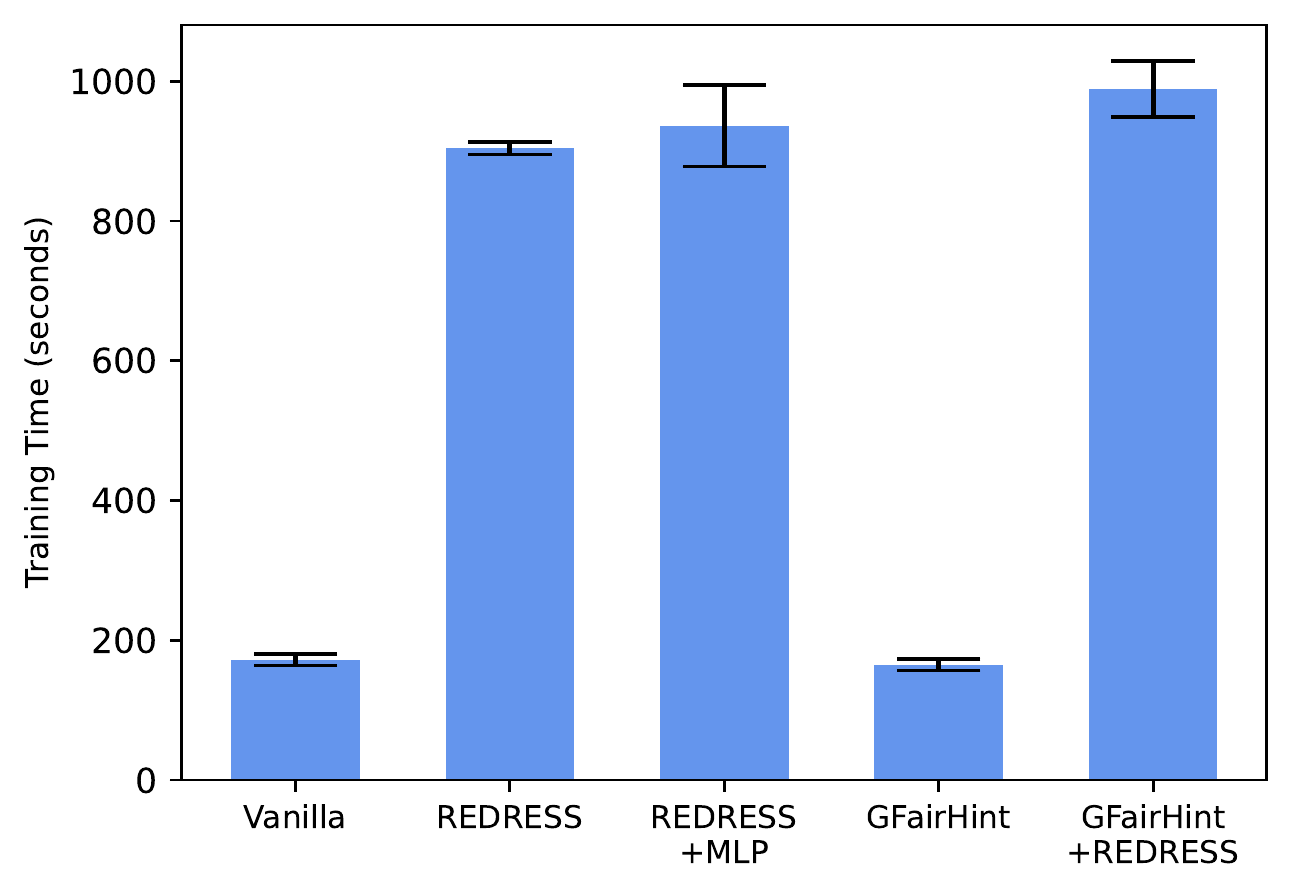}
    \caption{Total time of different models for training 50 epochs.}
    \label{fig:training_time}
\end{figure}

To show the gap more clearly, we perform efficiency evaluation experiments on the largest dataset, ArXiv, which has 90k training nodes. 
We choose GCN as the backbone model. 
The experiments were conducted in a controlled computation environment with single GPU (RTX2080ti) and fixed GPU memory (12GB). 
For each method, we train the models for 300 epochs and visualize the average training time of 50 epochs in Figure \ref{fig:training_time}.

\section{Sensitivity Analysis on $K$ Value}
\label{sec:k_values}
\begin{figure}[t]
\centering
    \includegraphics[width=\linewidth]{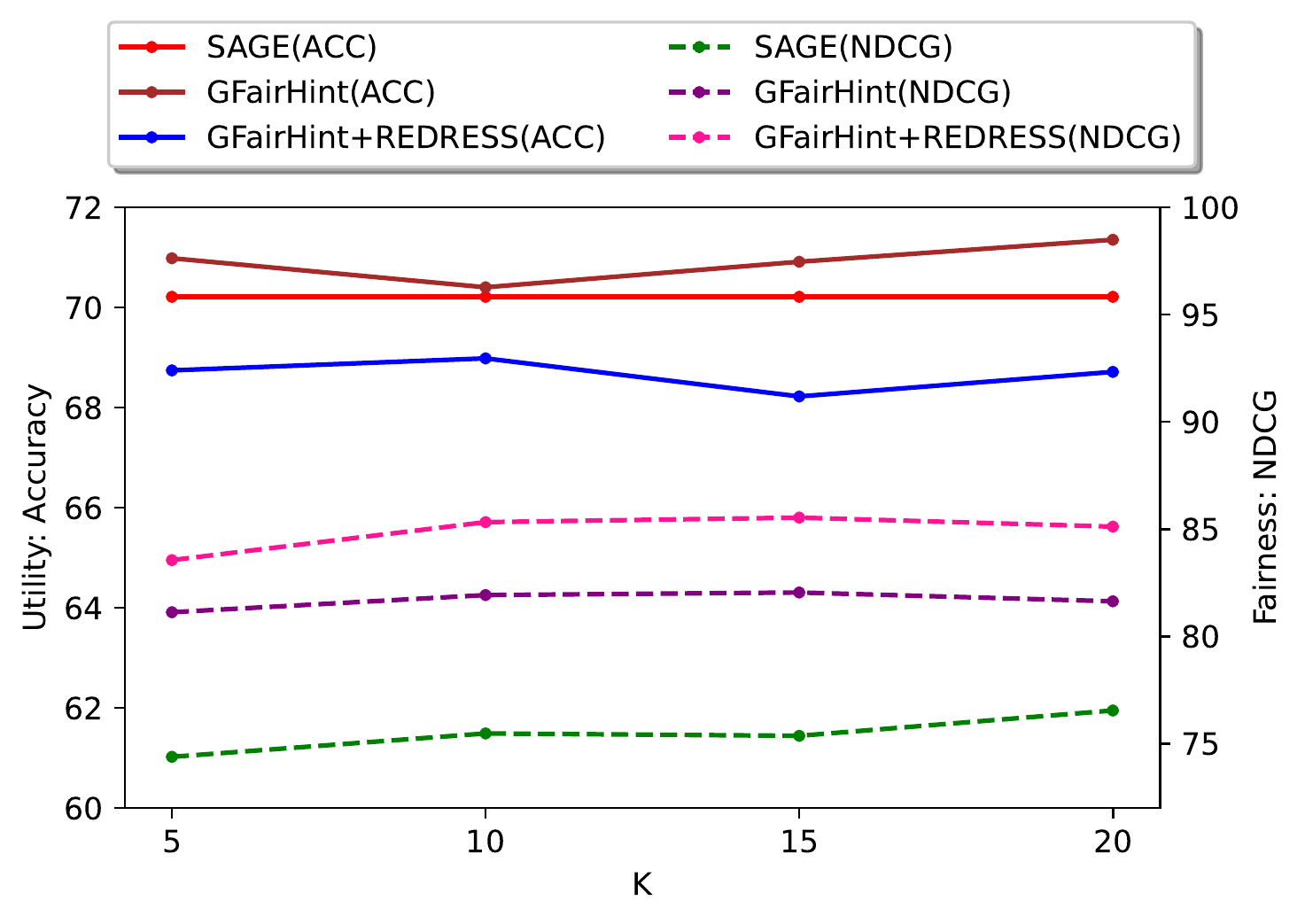}
    \caption{Influence of $k$ choices on Arxiv dataset across various methods. Our proposed method consistently outperforms baseline methods with varying $k$ values.}
    \label{fig:k_values}
\end{figure}

In this section, we investigate how the value of $k$, which is used to as the threshold to construct the fairness graph, will influence the performance of our proposed methods on both utility and individual fairness. We vary the $k$ among $[5, 10, 15, 20]$, reconstruct the fairness graph by connecting each node to its top-$k$ similar nodes, and regenerate the fairness hints. We reproduce the experiments with different $k$ values for Vanilla, GFairHint, and GFairHint + REDRESS methods on the Arxiv dataset with SAGE as the backbone model. 
Figure \ref{fig:k_values} shows the accuracy and fairness (NDCG@K) with varying values of $k$. From the figure, we can observe that Our proposed GFairHint and GFairHint + REDRESS methods consistently outperform the baseline methods with multiple values of $k$. The utility of our methods is minimally affected as the value of k increases. This pattern indicates that GFairHint successfully strikes a suitable balance between preserving model utility and promoting individual fairness across different thresholds of fairness graph construction.
\end{document}